\documentclass[fleqn,10pt]{wlscirep}
\usepackage[utf8]{inputenc}
\usepackage[T1]{fontenc}
\usepackage{xspace}
\usepackage{multirow}

\newcommand{\projectname}{\textbf{StepFM}\xspace}

\title{Physical activities enable scalable foundation modelling for broad-spectrum health prediction}

\author[1]{Zhenghuang Wu}
\author[2]{Yuyao Zhu}
\author[3,*]{Songlin Xu}

\affil[1]{Beihang University, Beijing, China}
\affil[2]{University of Science and Technology of China, Hefei, China}
\affil[3]{The Hong Kong Polytechnic University, Hong Kong SAR, China}
\affil[*]{corresponding.songlin.xu@polyu.edu.hk}

\begin{abstract}
Wearable and mobile sensing technologies have demonstrated strong potential for health inference; however, most sensor models are designed for specific disease types, limiting their transferability across different health risks. Wearable foundation models  offer a more generalizable approach in diverse health risk types. Nevertheless, most existing methods rely on high-frequency raw sensor data, raising concerns about privacy, computational overhead, and scalability across devices and populations. In this paper, we propose \projectname, a foundation model built solely on step counter data for broad-spectrum health prediction. Leveraging the ubiquity and low-dimensional nature of step data, \projectname provides a practical, privacy-preserving, and computation-efficient alternative to traditional sensor-based models. We design a scalable pre-training framework that captures temporal dynamics and behavioral patterns from large-scale step sequences, enabling transfer across more than 20 health risk prediction tasks spanning diverse devices, new regions, and novel disease types. Extensive experiments demonstrate that \projectname achieves strong performance compared to existing methods while maintaining robustness across heterogeneous settings. Furthermore, our analysis reveals interpretable and generalizable relationships between physical activity patterns and various health risks, offering new insights into activity-based health modeling. Our work establishes step-based sensing as a viable foundation for scalable and real-world health monitoring.
\end{abstract}

\begin{document}

\flushbottom
\maketitle

\section*{Introduction}

The rapid proliferation of wearable and mobile devices has enabled continuous monitoring of human health at an unprecedented scale \cite{zhu2025master,petropoulos2020wearable, kellner2017online, bisio2024towards,li2025sensorllm,luo2025toward,miao2026wonderwall}. Among various sensing modalities, motion signals have emerged as a powerful substrate to infer a wide spectrum of health-related outcomes, ranging from chronic disease progression \cite{farooqi2015predicting} to mental health \cite{ruan2025foundation}. Wearable sensing technologies, particularly inertial measurement units (IMUs), have been widely adopted to quantify human motion and support clinical assessment, rehabilitation, and disease monitoring \cite{petropoulos2020wearable, kellner2017online, bisio2024towards}. Prior studies have demonstrated that motion-derived features such as gait and posture are strongly correlated with health risks, including fall risk and functional decline \cite{buckinx2015added, embry2025strategic, nguyen2018fall}, and can serve as biomarkers for neurological disorders such as Parkinson’s disease \cite{an2021mgait, zhang2019pdmove}. These findings highlight the feasibility of motion sensing as a scalable and non-invasive approach for health inference.

Recent advances further push this paradigm toward large-scale learning, where foundation models trained on massive sensor datasets demonstrate strong generalization and transferability across tasks and populations \cite{abbaspourazad2023large, narayanswamy2024scaling, gu2025sensing, erturk2025beyond}. These developments suggest a promising direction in which a single model can support diverse health inference tasks in a unified and scalable manner \cite{ruan2025foundation}.

Despite this progress, existing approaches predominantly rely on high-frequency raw sensor data such as IMU signals or physiological waveforms \cite{zhu2025master,li2025sensorllm,luo2025toward,miao2026wonderwall}. While informative, such data introduces several practical challenges. First, raw motion signals encode fine-grained behavioral patterns that may raise privacy concerns \cite{10.1145/3658644.3690370}, limiting their deployment in real-world settings. Second, processing high-frequency time series requires substantial computational resources, particularly for long-term and population-scale monitoring \cite{narayanswamy2024scaling}. Third, models trained on raw sensor data are often tightly coupled with specific devices, sensor configurations, and sampling rates, which hinders their ability to generalize across datasets and populations \cite{qiu2025towards}. These limitations highlight a fundamental trade-off between sensing fidelity and scalability in wearable health systems.

In contrast, step counts, readily available from commodity pedometers and smartphones, offer a compelling alternative. As a compact and low-dimensional representation of human activity, step data is inherently privacy-preserving, computationally efficient, and widely accessible. Prior studies have demonstrated that step-based measures are strongly associated with diverse health outcomes, including mental health conditions, chronic diseases, recovery trajectories, and mortality risk \cite{fujino2023decreased, hafiz2020wearable, farooqi2015predicting, bassett2010pedometer}. At the population level, step counts have been linked to energy expenditure, cardiometabolic health, and overall well-being \cite{tudor2004descriptive, eisenmann2007utility}, further supporting their ecological validity and scalability .

However, most existing studies \cite{hossain2025predicting,farooqi2015predicting} treat step data as a task-specific signal, designing separate models for individual health outcomes. This fragmented paradigm limits the ability to capture shared structure across heterogeneous health risks and prevents step data from being used as a unified representation for broad health inference.

To address this gap, we propose \textbf{\projectname}, a foundation model built solely on step count data for broad-spectrum health prediction. Instead of relying on raw motion signals, \projectname leverages large-scale step sequences to learn generalizable representations of human activity, enabling unified modeling across diverse health outcomes. Our approach rethinks the role of low-dimensional behavioral signals in foundation modeling, demonstrating that compact and privacy-preserving data can still support rich and transferable representations.

We design a scalable pre-training framework that captures temporal dynamics and behavioral patterns from step data, and evaluate its effectiveness across more than 20 health risk prediction tasks spanning multiple public devices, new regions, and novel disease types. Beyond performance improvements, our experiments reveal consistent and interpretable relationships between physical activity patterns and diverse health risks, offering new insights into how daily behavior encodes health status at scale.

By bridging the gap between practicality and scalability, our work establishes step-based sensing as a viable foundation for next-generation health models, opening new opportunities for accessible and large-scale health monitoring in real-world settings.
In summary, this paper makes the following contributions:
\begin{itemize}
    \item Pre-trained on 141.12 million minute-level step-count observations, we develop a foundation model with 3.4 million trainable parameters that relies solely on step-count data to predict a broad spectrum of health risks, offering a practical, privacy-preserving, and computationally efficient alternative to raw sensor–based approaches.
    \item We design a scalable pre-training framework that supports transfer across more than 20 health prediction tasks, spanning diverse devices, new regions, and novel diseases.
    \item We conduct extensive experiments that demonstrate strong performance and uncover generalizable insights into the relationship between physical activity and health risks, providing both empirical validation and scientific interpretability.
\end{itemize}

\section*{Results}

\subsection*{Experimental Settings}

To evaluate whether \projectname can serve as a scalable step-based foundation model for broad-spectrum health prediction, we compare it with representative baselines from related areas of wearable health modeling, including general wearable foundation models, physiological foundation models, actigraphy foundation models, step-count representation learning, and generic time-series pre-training. In addition, we include a traditional machine learning baseline based on hand-crafted step-count features as an interpretable reference point. This baseline is not intended as a foundation model, but helps assess whether learned representations capture and extend known activity-derived health signals.

\begin{itemize}
    \item \textbf{NormWear \cite{luo2025toward}.} NormWear is a wearable sensing foundation model for heterogeneous physiological signals, including PPG, ECG, EEG, GSR, and IMU. It uses multi-scale signal representations and channel-aware attention to support different sensor configurations and health-related downstream tasks.
    \item \textbf{Pulse-PPG \cite{saha2025pulse}.} Pulse-PPG is an open-source PPG foundation model trained on field-collected wearable PPG data. It uses motif-based relative contrastive learning to learn robust temporal representations from noisy real-world physiological signals.
    \item \textbf{PAT \cite{ruan2025foundation}.} The Pretrained Actigraphy Transformer is a foundation model for wearable movement data. It uses patch embeddings and masked reconstruction pretraining on large-scale NHANES actigraphy data, and is evaluated on mental-health-related tasks such as depression, sleep disorders, and medication usage.
    \item \textbf{Hossain et al.’s Self-supervised Contrastive Pretraining Approach (SSCP) \cite{hossain2025predicting}.} Hossain et al. proposed a self-supervised contrastive pretraining approach using daily step count sequences to predict depression treatment improvement. The approach learns step count representations through contrastive pretraining and then uses the learned representations for supervised classification.
    \item \textbf{TimeSiam \cite{dong2024timesiam}.} TimeSiam is a self-supervised pre-training framework that uses Siamese encoders to model temporal correlations through a past-to-current reconstruction task with simple masked augmentation. It learns time-dependent representations from randomly sampled past and current subseries, further using lineage embeddings to distinguish temporal distances. TimeSiam achieves strong forecasting and classification performance across 13 standard benchmarks in both intra- and cross-domain scenarios.
    \item \textbf{Trad. ML \cite{lavalley2008logistic}.} Trad. ML is a logistic regression (LR) baseline built on hand-crafted step-count features, including hourly, daily and weekly activity statistics, sedentary patterns, circadian regularity, and bout-level activity descriptors. We include it as an interpretable reference baseline to contextualize the gains of learned representations over established activity-derived health indicators.
\end{itemize}

\begin{table*}[t]
    \centering
    \caption{\textbf{Detailed performance comparison (AUROC) on 21 downstream health risk prediction tasks.} The best performance for each task is highlighted in \textbf{bold}, and the second-best performance is underlined.}
    \label{tab:comparison_auroc}
    \resizebox{\textwidth}{!}{
    \begin{tabular}{l c c c c c c c}
        \toprule
        \textbf{Downstream Tasks} & Pulse-PPG & PAT & SSCP & TimeSiam & NormWear & \textbf{\projectname} & \textit{Trad. ML} \\
        \midrule
        \multicolumn{8}{l}{\textit{Cardiovascular \& Circulatory}} \\
        \midrule
        Angina & 0.5991 & 0.7376 & 0.6829 & 0.6566 & \textbf{0.7565} & \underline{0.7524} & 0.7398 \\
        Heart Failure & 0.6201 & 0.7905 & 0.7513 & 0.7440 & \underline{0.8089} & \textbf{0.8320} & 0.8065 \\
        Coronary Disease & 0.6098 & 0.7253 & 0.7467 & 0.7206 & \underline{0.7758} & \textbf{0.7846} & 0.7748 \\
        Heart Attack & 0.6249 & 0.7367 & 0.7273 & 0.7450 & \underline{0.8092} & \textbf{0.8196} & 0.8041 \\
        Stroke & 0.6435 & 0.7443 & 0.7713 & 0.7446 & \underline{0.7985} & \textbf{0.8284} & 0.7851 \\
        High Blood Pressure & 0.6688 & 0.3766 & 0.6558 & 0.6893 & 0.7157 & \textbf{0.7544} & \underline{0.7424} \\
        \midrule
        \multicolumn{8}{l}{\textit{Metabolic, Endocrine \& Renal}} \\
        \midrule
        Diabetes & 0.6843 & 0.7207 & 0.6863 & 0.7672 & 0.8072 & \textbf{0.8509} & \underline{0.8293} \\
        Overweight & 0.5572 & 0.5521 & 0.5647 & 0.6098 & 0.6214 & \textbf{0.6708} & \underline{0.6511} \\
        High Cholesterol & 0.6026 & 0.5166 & 0.6079 & 0.6168 & 0.6366 & \textbf{0.6789} & \underline{0.6459} \\
        Gout & 0.5831 & 0.6276 & 0.6983 & 0.6427 & 0.7095 & \textbf{0.7502} & \underline{0.7274} \\
        Thyroid Problem & 0.5536 & 0.6357 & 0.6186 & 0.6095 & \underline{0.6372} & \textbf{0.6517} & 0.6224 \\
        Kidney Weak/Failing & 0.5255 & 0.6624 & 0.6846 & 0.7030 & \underline{0.7031} & \textbf{0.7153} & 0.7014 \\
        \midrule
        \multicolumn{8}{l}{\textit{Respiratory, Immune \& Oncology}} \\
        \midrule
        Emphysema & 0.6576 & 0.7801 & \underline{0.7883} & 0.7347 & 0.7795 & \textbf{0.7988} & 0.7305 \\
        Chronic Bronchitis & 0.5461 & 0.6050 & 0.6151 & 0.5924 & \underline{0.6373} & \textbf{0.6489} & 0.6012 \\
        Arthritis & 0.6468 & 0.6294 & 0.6482 & 0.6622 & 0.7050 & \textbf{0.7335} & \underline{0.7294} \\
        Cancer & 0.6133 & 0.6234 & 0.6429 & 0.6602 & 0.7062 & \textbf{0.7432} & \underline{0.7319} \\
        Anemia Treatment & 0.5278 & 0.5666 & 0.5775 & 0.6120 & \underline{0.6448} & \textbf{0.6597} & 0.6287 \\
        \midrule
        \multicolumn{8}{l}{\textit{Neurological, Mental \& Sensory}} \\
        \midrule
        Memory-Cognitive & 0.5160 & 0.5243 & 0.5811 & \underline{0.6566} & 0.6474 & \textbf{0.6571} & 0.6222 \\
        Depression & 0.5575 & 0.5958 & 0.5626 & 0.6445 & \underline{0.6499} & \textbf{0.6642} & 0.6195 \\
        Sleep Disorder & 0.5767 & 0.5389 & 0.5940 & 0.6331 & 0.6639 & \textbf{0.7014} & \underline{0.6788} \\
        Vision Trouble & 0.5751 & 0.5706 & 0.5840 & 0.6335 & 0.6532 & \textbf{0.6709} & \underline{0.6643} \\
        \midrule
        \textbf{Mean AUROC} & 0.5947 & 0.6314 & 0.6566 & 0.6704 & \underline{0.7079} & \textbf{0.7318} & 0.7065 \\
        \bottomrule
    \end{tabular}
    }
\end{table*}

\subsection*{Comparison with Existing Foundation Models}
\label{sec:comparison}

To evaluate the effectiveness and generalizability of the proposed framework, we compare \projectname against a set of baselines, including a generic time-series pre-training method (TimeSiam), modality-specific or general wearable foundation models (Pulse-PPG, PAT, NormWear, and SSCP), and a traditional machine learning baseline. The evaluation is conducted across 21 downstream health prediction tasks using the full-shot linear probing protocol and AUROC as the primary metric.

The detailed performance comparison is summarized in Table~\ref{tab:comparison_auroc} and Figure~\ref{fig:disease_trend} illustrates the overall performance trends across diseases. Overall, our model achieves the best mean AUROC of 0.7318 and ranks first in 20 out of 21 tasks. 

\textbf{Generic Time-Series Pre-training Model:} TimeSiam provides a general-purpose self-supervised baseline for temporal representation learning, but it is not specifically designed for sparse, low-dimensional, and highly periodic step-count sequences. In Table~\ref{tab:comparison_auroc}, TimeSiam reaches a mean AUROC of 0.6704, whereas our method improves it to 0.7318, corresponding to a relative gain of 9.2\%. This suggests that generic reconstruction objectives alone are insufficient to capture long-term behavioral phenotypes for broad-spectrum health prediction.

\textbf{Modality-Specific and General Wearable Foundation Models:} Pulse-PPG, PAT, SSCP, and NormWear represent foundation-model baselines trained on different wearable or step-count signals. Pulse-PPG, PAT, and SSCP underperform in this broad-spectrum setting, reflecting the difficulty of transferring representations learned from high-frequency PPG, generic actigraphy, or depression-oriented step-count pre-training to diverse health prediction tasks. NormWear is the strongest foundation-model baseline, with a mean AUROC of 0.7079, and it marginally outperforms \projectname on Angina. However, our model maintains a clear advantage in all remaining tasks. The improvements are especially visible for chronic conditions such as Diabetes, Heart Failure, and Stroke, where behavior-aware temporal modeling provides stronger health-relevant representations than generic multimodal wearable pre-training.

\textbf{Traditional Machine Learning Baseline:} The traditional ML baseline is also highly competitive, achieving a mean AUROC of 0.7065 and outperforming several deep baselines. This indicates that carefully engineered step-derived statistics already encode meaningful health signals and provide a strong reference point for representation learning. As shown in Figure~\ref{fig:disease_trend}, different models follow similar disease-level performance trends, while our model consistently raises the upper bound of these trends. Together, these results suggest that our method extends beyond manually engineered activity indicators while preserving their clinically relevant structure.

\subsection*{Ablation on Model Modules}
\label{sec:ablation_modules}

We conduct an ablation study to examine how each major component contributes to downstream disease prediction. Starting from a vanilla Mamba backbone, we incrementally add the proposed tokenization and temporal encoding, the micro-stream FiLM module, and the activity phenotype alignment objective. Table~\ref{tab:ablation} reports the mean AUROC across all 21 health outcomes. We use AUROC as the primary ablation metric because the downstream labels are highly imbalanced and AUROC is more robust to such class skew than threshold-dependent metrics.

\textbf{Native Mamba Baseline:} 
The baseline uses a vanilla Mamba backbone trained with autoregressive next-token prediction on hourly step-count sequences, without log-scaled tokenization, temporal rhythm encoding, or phenotype alignment. The selective state-space design of Mamba is well suited to multi-day step sequences, as it efficiently captures long-range temporal dependencies, and the next-token objective directly forces the model to model sequential transitions in step counts. This configuration already achieves a mean AUROC of 0.7007, indicating that sequence-level modeling of step counts alone is informative for health prediction.

\textbf{Effect of Tokenization and Temporal Rhythm Encoding:} 
Hourly step counts are highly skewed and largely sparse, so a naive linear discretization wastes vocabulary on rare large values and loses resolution in the low-to-moderate activity range. We therefore replace the naive input representation with a log-scaled tokenizer that allocates higher resolution to clinically meaningful low and moderate activity levels. In addition, the same step volume can carry very different behavioral meaning depending on time of day and day of week (e.g., walking at noon versus at 3:00 AM). To make this context explicit, we inject deterministic Fourier-based temporal rhythm encoding for diurnal and weekly periodicity. With both components, the mean AUROC improves to 0.7095, confirming that step-volume discretization and temporal grounding provide complementary signal beyond raw sequence modeling.

\textbf{Effect of Micro Stream with FiLM:} 
Hourly step totals aggregate heterogeneous within-hour activity morphologies into a single scalar value. As a result, behaviors with very different physiological implications, such as a sustained 30-minute jog and sporadic household movements, can yield identical hourly sums. Such micro-level dynamics are clinically informative yet remain invisible to a purely macro hourly stream. To recover them, we introduce a 1D convolutional micro stream over the minute-level waveform within each hour, and employ FiLM to dynamically condition the macro Mamba hidden states on these local features. This design allows minute-level morphology to modulate the macro representation without disrupting its long-range autoregressive flow. As a result, the mean AUROC increases to 0.7282, demonstrating that fine-grained local dynamics provide a strong complement to hourly macro modeling.

\textbf{Full \projectname Framework:} 
Optimizing only for next-token prediction can bias the model toward short-horizon transitions and under-represent macroscopic behavioral phenotypes such as daily activity volume, sedentary burden, and cross-day regularity. To explicitly encode these clinically meaningful indicators, the full framework adds hierarchical activity phenotype alignment at hourly, daily, and weekly scales during pre-training. The full model achieves the best overall performance, with a mean AUROC of 0.7318.

\begin{table}[t]
    \centering
    \caption{\textbf{Ablation study on the core modules of \projectname.} The table shows the incremental performance gains from log-scaled tokenization with temporal rhythm encoding, micro-stream FiLM modulation, and activity phenotype alignment. The best performance for each task is highlighted in \textbf{bold}, and the second-best performance is underlined.}
    \label{tab:ablation}
    \begin{tabular}{l c c c c c}
        \toprule
        \textbf{Model Variant} & \textbf{Tokenization \& Time} & \textbf{Micro Stream} & \textbf{Activity Stats} & \textbf{Mean AUROC} &  \\
        \midrule
        Native Mamba          &            &            &            & 0.7007  \\
        + Log \& Time         & \checkmark &            &            & 0.7095 \\
        + Micro FiLM          & \checkmark & \checkmark &            & \underline{0.7282} \\
        \textbf{Full \projectname} & \checkmark & \checkmark & \checkmark & \textbf{0.7318} \\
        \bottomrule
    \end{tabular}
\end{table}

\subsection*{\projectname Captures cross-Disease Correlation}
\label{sec:cross_disease_correlation}

\begin{figure*}[t]
    \centering
    \includegraphics[width=\textwidth]{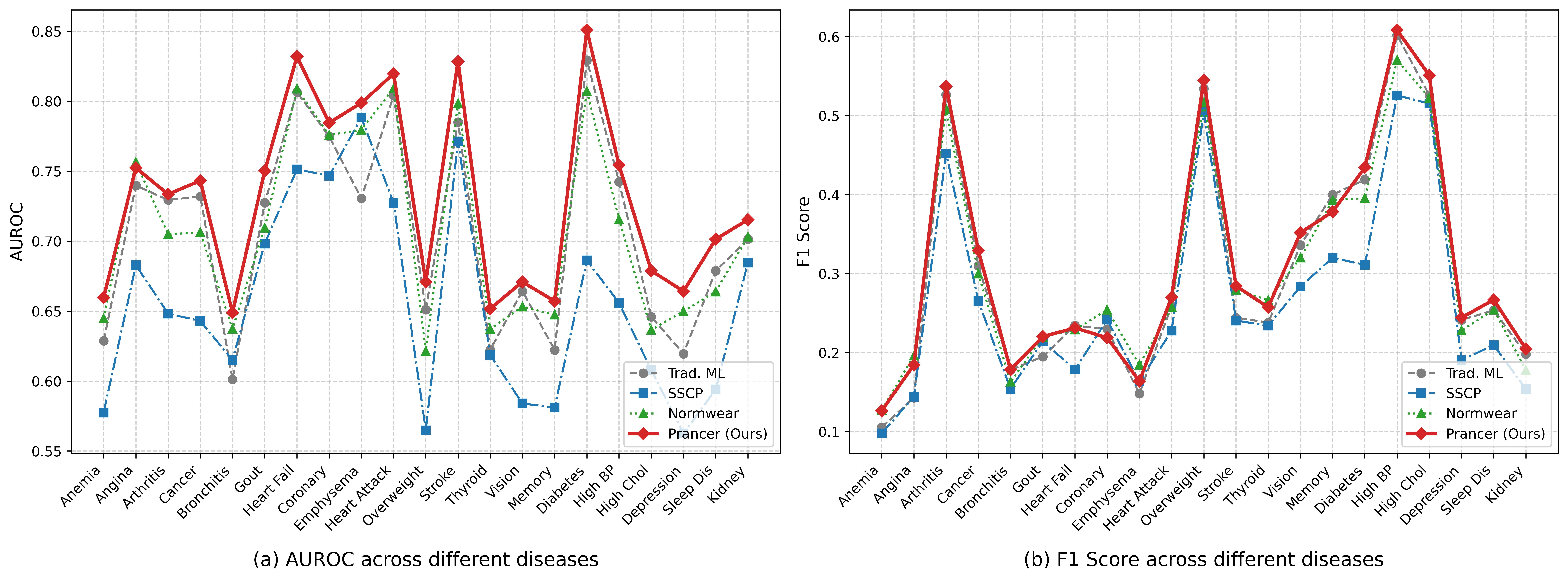} 
    \caption{\textbf{Performance trends across 21 diseases.} (a) AUROC scores and (b) F1 scores for different models. The parallel oscillation across models indicates that disease predictability is primarily driven by the inherent correlation between the specific disease and physical activity patterns.}
    \label{fig:disease_trend}
\end{figure*}

To further elucidate the relationship between step-count sequences and diverse health outcomes, we analyzed the performance trajectories across all 21 evaluated diseases. As illustrated in Figure \ref{fig:disease_trend}, we plot the AUROC and F1 scores of four representative models across various disease endpoints, including a traditional machine learning baseline, a step-count-based foundation model (SSCP), the strongest wearable foundation model baseline (NormWear), and our proposed \projectname.

A salient observation from these visualizations is the remarkable morphological consistency of the performance curves across all evaluated models. Despite substantial differences in architectural complexity and absolute predictive capability, the models exhibit parallel oscillations. Specifically, all models consistently achieve peak AUROC and F1 scores for conditions intrinsically linked to mobility and cardiometabolic health, such as Diabetes, Heart Failure, Heart Attack, Stroke, and High Blood Pressure. Conversely, predictive performance uniformly diminishes across all methods for conditions such as Anemia, Chronic Bronchitis, and Depression.

This synchronized variability strongly implies that the predictability of a given health risk is fundamentally constrained by its inherent biological and behavioral correlation with physical activity, rather than the choice of modeling architecture. It corroborates the premise that step counts serve as a highly sensitive digital biomarker for cardiovascular and metabolic conditions, consistent with prior epidemiological evidence linking daily step counts to lower risks of mortality, cardiovascular disease, and dysglycemia~\cite{hall2020systematic}. In contrast, step counts exhibit a naturally lower signal-to-noise ratio for certain respiratory and mental health disorders, where the relationship with physical activity is weaker, more indirect, or modulated by additional clinical and contextual factors~\cite{yuan2025association,rice2020reduced}.

Notably, the absolute F1 scores for several conditions remain relatively low across all models. As shown in Figure~\ref{fig:disease}, this phenomenon is primarily driven by severe class imbalance in the dataset. Many conditions exhibit an extreme long-tail distribution, in which positive cases account for only a small fraction of the population (e.g., $1.0\%$ for Emphysema and $1.4\%$ for Angina). As the F1 score is highly sensitive to such skewed distributions and heavily penalizes false positives in the presence of a dominant negative class, its absolute value is naturally suppressed. Even under these inherent modality and data constraints, \projectname consistently achieves the strongest empirical performance, in both AUROC and F1, across the entire disease spectrum. This sustained superiority demonstrates that \projectname effectively captures and amplifies the underlying cross-disease correlations, learning robust and epidemiologically grounded representations rather than relying on dataset-specific artifacts.

\subsection*{Why Does \projectname Work?}

The traditional machine learning baseline is built on 99 hand-crafted step-count features, including circadian rest-activity rhythm descriptors~\cite{witting1990alterations}, daily and weekly step volume statistics~\cite{tudor2004many}, sedentary and active bout morphology, and population-level actigraphy summaries widely used in large-scale physical activity studies~\cite{doherty2017large}. These features are grounded in established epidemiological evidence that daily step volume, activity intensity, and circadian regularity are sensitive behavioral markers of cardiovascular, metabolic, and mental health conditions~\cite{hall2020systematic,bizzozero2024daily,pearce2022association}. The fact that such a transparent linear model already attains a mean AUROC of 0.7065 confirms that step-derived behavioral signals carry substantial health information, and provides a strong reference point for evaluating learned representations.

Building on this baseline, our method achieves a higher mean AUROC of 0.7318 and consistently improves F1 across the disease spectrum. We attribute this improvement to two complementary capabilities. First, the behavior-aware dual-stream architecture natively captures the same interpretable spatio-temporal dynamics that drive the success of hand-crafted features, including circadian rhythms, multi-scale activity volume, and active/sedentary patterns, without requiring explicit feature engineering. Second, the model further extracts latent behavioral patterns that are difficult to specify manually, such as long-range cross-day regularities, micro-level intra-hour dynamics, and contextual interactions between time of day and activity intensity. By unifying these explicit and latent behavioral cues, our model ensures that its predictions remain grounded in observable, clinically meaningful human behaviors while pushing predictive performance beyond what manually engineered features alone can achieve.

\subsection*{Few-Shot to Full-Shot Performance}
\label{sec:scaling_laws}

A fundamental characteristic of an effective foundation model is its compatibility with scaling laws, demonstrating robust generalization under constrained conditions. We evaluate \projectname's scaling behaviors across two critical dimensions: the size of the downstream labeled dataset and the length of the input observation window. In addition to the overall mean performance, we further track two representative conditions, Gout and Depression, to verify that the observed scaling behaviors are not driven solely by averaging effects but also hold for individual disease prediction tasks.

\begin{figure}[t]
    \centering
    \includegraphics[width=\linewidth]{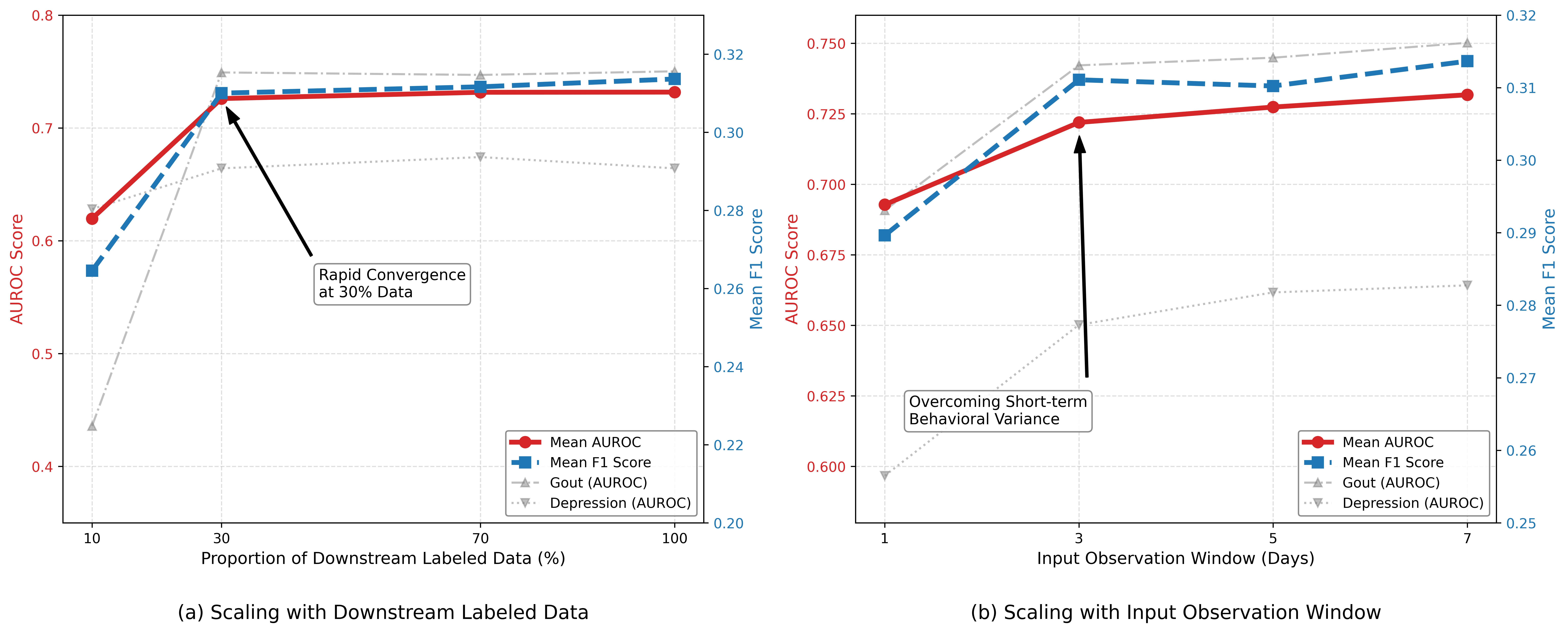}
    \caption{\textbf{Scaling behaviors of \projectname.} (a) Data Efficiency: performance under varying proportions of downstream labeled training data. (b) Temporal Receptive Field: performance under varying lengths of the input observation window.}
    \label{fig:finetune_scaling_law}
\end{figure}

\textbf{Scaling with Downstream Labeled Data.} 
In real-world clinical applications, acquiring large-scale, high-quality disease annotations is notoriously expensive and labor-intensive. To evaluate the downstream data efficiency, we systematically vary the proportion of the downstream training set from 10\% to 100\% while keeping the test set fixed. As illustrated in Figure~\ref{fig:finetune_scaling_law}(a), our method exhibits a classic logarithmic learning curve. Using only 30\% of the labeled data, the model already reaches a mean AUROC of 0.7260, recovering more than 95\% of its full-shot performance. A consistent trend is observed at the individual disease level, where the AUROC for Gout improves from 0.4358 to 0.7492 as the labeled data increases from 10\% to 100\%. These results suggest that the behavior-aware pre-training embeds highly transferable physiological representations that require minimal task-specific supervision for effective adaptation, making \projectname particularly advantageous for modeling rare conditions or populations with limited annotated cohorts.

\textbf{Scaling with Input Observation Window.} 
Beyond annotated data size, the predictive power of behavioral biomarkers inherently depends on the duration of the observation window. Short-term activity snapshots are easily distorted by transient behavioral anomalies, such as temporary illnesses, vacations, or atypical work schedules, which can obscure an individual's true physiological baseline. To evaluate \projectname's temporal receptive field, we scale the input step-count sequence length from 1 day to 7 days. As illustrated in Figure~\ref{fig:finetune_scaling_law}(b), extending the observation window yields a consistent and monotonic improvement in predictive accuracy. With only 1 day of data, our model reaches a mean AUROC of 0.6928, providing a basic predictive foundation. Extending to 3 days raises the mean AUROC to 0.7220, indicating that the model successfully abstracts stable cross-day regularities to overcome short-term variance, after which performance continues to climb and gradually plateaus by the 7-day mark. This trend is especially pronounced for individual conditions such as Depression, where the AUROC increases from 0.5967 with 1 day of data to 0.6642 with 7 days. Overall, these results validate our hypothesis that long-term macroscopic rhythms construct a more comprehensive and noise-resilient physiological phenotype than isolated short-term snapshots, and they further demonstrate the efficacy of our dual-stream architecture in handling ultra-long temporal contexts.

\subsection*{Layer-wise Representation Analysis}
\label{sec:layer_ablation}

\begin{figure}[t]
    \centering
    \includegraphics[width=0.5\linewidth]{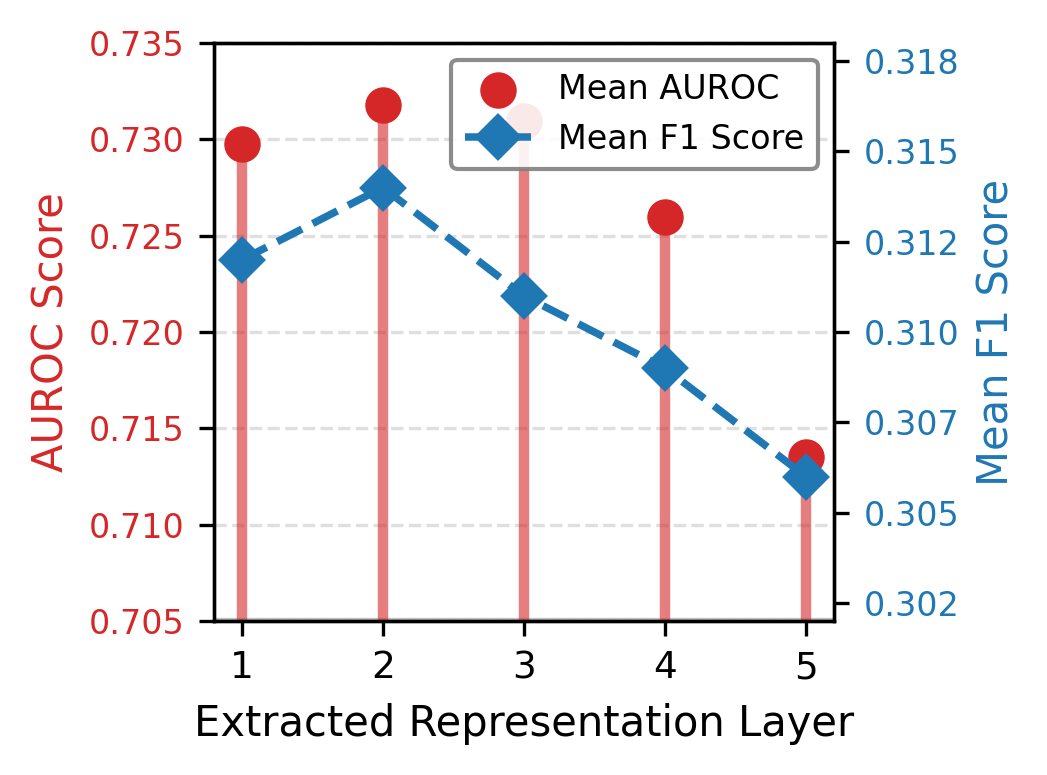} 
    \caption{\textbf{Linear probing performance using representations extracted from different Mamba layers.} The inverted-U trajectory indicates that intermediate layers retain optimal, generalizable physiological representations.}
    \label{fig:layer_ablation}
\end{figure}

To understand the hierarchical feature learning dynamics of our dual-stream Step-Mamba encoder, we investigate the downstream linear probing performance using representations extracted from Layer 1 to Layer 5. As illustrated in Figure~\ref{fig:layer_ablation}, performance does not monotonically increase with layer depth, in contrast to the common observation in fully supervised models. Instead, it follows a clear inverted-U trajectory, peaking at Layer 2 with a mean AUROC of 0.7318 and gradually declining to 0.7135 at Layer 5. The mean F1 score exhibits the same inverted-U trend, reaching its maximum of 0.3137 at Layer 2 before decreasing in deeper layers.

This phenomenon reveals a misalignment between the pre-training objectives and downstream clinical tasks. Since the pre-training loss is dominated by autoregressive next-token prediction and explicit activity-statistics alignment, the topmost layers become increasingly specialized in forecasting short-term step transitions, discarding generalized physiological cues that are not directly useful for next-step prediction. The lower-to-middle layers, in contrast, retain a more balanced representation of micro-level activity morphology and macro-level temporal rhythms, which is better suited to broad-spectrum disease prediction. Extracting representations from these intermediate layers therefore provides a stronger inductive bias for downstream clinical inference and mitigates the gap between self-supervised pre-training and clinical tasks.

\begin{table}[htbp]
\centering
\caption{\textbf{Per-task AUROC on NHANES 2005--2006 with the full downstream training set.} The best performance for each task is highlighted in \textbf{bold}, and the second-best performance is underlined.}
\label{tab:detailed_full}
\resizebox{\textwidth}{!}{
\begin{tabular}{l c c c c c c}
    \toprule
    \textbf{Downstream Tasks} & Pulse-PPG & PAT & SSCP & TimeSiam & NormWear & \textbf{\projectname} \\
    \midrule
    \multicolumn{7}{l}{\textit{Cardiovascular \& Circulatory}} \\
    \midrule
    Angina & 0.4903 & 0.5086 & \underline{0.6759} & 0.6055 & 0.6725 & \textbf{0.6988} \\
    Heart Failure & 0.5451 & 0.5433 & 0.7078 & 0.6747 & \underline{0.7121} & \textbf{0.7345} \\
    Coronary Disease & 0.4923 & 0.5277 & \underline{0.6773} & 0.6512 & 0.6371 & \textbf{0.7013} \\
    Heart Attack & 0.5943 & 0.6368 & 0.7128 & 0.7201 & \underline{0.7586} & \textbf{0.7628} \\
    Stroke & 0.5209 & \underline{0.7426} & 0.7165 & 0.6832 & 0.7313 & \textbf{0.7756} \\
    High Blood Pressure & 0.6085 & 0.6066 & 0.6474 & 0.6760 & \underline{0.7188} & \textbf{0.7418} \\
    \midrule
    \multicolumn{7}{l}{\textit{Metabolic, Endocrine \& Renal}} \\
    \midrule
    Diabetes & 0.6061 & 0.6709 & 0.6340 & \textbf{0.7594} & 0.7467 & \underline{0.7591} \\
    Overweight & 0.5385 & 0.4815 & 0.5681 & 0.5597 & \underline{0.5981} & \textbf{0.6218} \\
    High Cholesterol & 0.5200 & 0.4942 & 0.5454 & \underline{0.6039} & 0.5690 & \textbf{0.6069} \\
    Thyroid Problem & 0.5781 & 0.4941 & 0.6137 & \underline{0.6382} & 0.6330 & \textbf{0.6761} \\
    Kidney Weak/Failing & 0.5148 & 0.5785 & \underline{0.6280} & 0.5360 & 0.6121 & \textbf{0.6292} \\
    \midrule
    \multicolumn{7}{l}{\textit{Respiratory \& Systemic Conditions}} \\
    \midrule
    Emphysema & 0.6936 & 0.6719 & 0.4904 & 0.6755 & \underline{0.6978} & \textbf{0.7031} \\
    Chronic Bronchitis & 0.4889 & 0.5842 & 0.5271 & 0.5052 & \underline{0.6240} & \textbf{0.6461} \\
    Arthritis & 0.5551 & 0.6070 & 0.6320 & 0.6838 & \underline{0.6839} & \textbf{0.7026} \\
    Cancer & 0.5931 & 0.5141 & 0.6359 & \textbf{0.7432} & \underline{0.7016} & 0.6936 \\
    Anemia Treatment & 0.6082 & 0.5802 & \underline{0.6852} & 0.6729 & \textbf{0.7002} & 0.6670 \\
    \midrule
    \multicolumn{7}{l}{\textit{Neurological, Mental \& Sensory}} \\
    \midrule
    Depression & 0.5932 & 0.5129 & 0.5842 & 0.6246 & \underline{0.6327} & \textbf{0.6602} \\
    Sleep Disorder & 0.5318 & 0.5067 & 0.5549 & \textbf{0.6150} & 0.5799 & \underline{0.5930} \\
    Vision Trouble & 0.5559 & 0.4830 & 0.5587 & 0.5758 & \underline{0.6057} & \textbf{0.6268} \\
    \midrule
    \textbf{Mean AUROC} & 0.5594 & 0.5655 & 0.6208 & 0.6423 & \underline{0.6640} & \textbf{0.6842} \\
    \bottomrule
\end{tabular}}
\end{table}

\begin{table}[htbp]
\centering
\caption{\textbf{Per-task AUROC and F1 on NHANES 2005--2006 under different downstream training-set sizes.} Mean AUROC and mean F1 are averaged over the 19 shared health outcomes and reported under both 30\% and full training data. The best performance for each task is highlighted in \textbf{bold}, and the second-best performance is underlined.}
\label{tab:summary_comparison}
\begin{tabular}{lcccc}
\toprule
\multirow{2}{*}{\textbf{Model}} & \multicolumn{2}{c}{\textbf{0.3}} & \multicolumn{2}{c}{\textbf{Full}} \\ \cmidrule(lr){2-3} \cmidrule(lr){4-5}
 & \textbf{AUROC} & \textbf{F1} & \textbf{AUROC} & \textbf{F1} \\ \midrule
Pulse-PPG & 0.5574 & 0.2136 & 0.5594 & 0.2153 \\
PAT & 0.5419 & 0.2124 & 0.5655 & 0.2105 \\
SSCP & 0.5915 & 0.2286 & 0.6208 & 0.2419 \\
TimeSiam & 0.6229 & 0.2282 & 0.6423 & 0.2461 \\
NormWear & \underline{0.6447} & \underline{0.2707} & \underline{0.6640} & \underline{0.2706} \\
\textbf{Prancer} & \textbf{0.6545} & \textbf{0.2710} & \textbf{0.6842} & \textbf{0.2867} \\ \bottomrule
\end{tabular}
\end{table}

\begin{table}[htbp]
\centering
\caption{\textbf{Per-task AUROC and F1 on the BarKA-MS dataset with the full downstream training set.} The two diseases are multiple-sclerosis-related and unseen during pre-training. The best performance for each task is highlighted in \textbf{bold}, and the second-best performance is underlined.}
\label{tab:barka_ms_full}
\begin{tabular}{llccccc}
\toprule
\textbf{Task} & \textbf{Metric} & \textbf{PAT} & \textbf{TimeSiam} & \textbf{NormWear} & \textbf{Pulse-PPG} & \textbf{Prancer} \\ \midrule
\multirow{2}{*}{Disability} 
& AUROC & 0.5561 & 0.6927 & 0.6923 & \underline{0.7990} & \textbf{0.8894} \\
& F1    & 0.8542 & 0.8454 & 0.8387 & \textbf{0.8866} & \underline{0.8817} \\ \midrule
\multirow{2}{*}{Fatigue}   
& AUROC & 0.4512 & 0.4094 & \underline{0.5786} & 0.5556 & \textbf{0.6638} \\
& F1    & 0.8542 & 0.8542 & \underline{0.8696} & \textbf{0.8911} & 0.8542 \\ \midrule
\multirow{2}{*}{Mean} 
& AUROC & 0.5037 & 0.5511 & 0.6354 & \underline{0.6773} & \textbf{0.7766} \\
& F1    & 0.8541 & 0.8497 & 0.8541 & \textbf{0.8888} & \underline{0.8679} \\ \bottomrule
\end{tabular}
\end{table}

\begin{table}[htbp]
\centering
\caption{\textbf{Per-task AUROC and F1 on the hour-level RESILIENT dataset with the full downstream training set.} As RESILIENT only provides hourly step counts, \projectname is evaluated without its minute-level micro-stream for a fair comparison with the hour-level baseline SSCP. The best performance for each task is highlighted in \textbf{bold}, and the second-best performance is underlined.}
\label{tab:resilient_full}
\begin{tabular}{llcc}
\toprule
\textbf{Task} & \textbf{Metric} & \textbf{SSCP} & \textbf{Prancer (w/o micro)} \\ \midrule
\multirow{2}{*}{Anxiety} 
& AUROC & \underline{0.6231} & \textbf{0.7890} \\
& F1    & \underline{0.2439} & \textbf{0.4000} \\ \midrule
\multirow{2}{*}{Cognitive Impairment}   
& AUROC & \underline{0.4739} & \textbf{0.4778} \\
& F1    & \underline{0.5233} & \textbf{0.5528} \\ \midrule
\multirow{2}{*}{Depression (PHQ)} 
& AUROC & \underline{0.7658} & \textbf{0.7864} \\
& F1    & \underline{0.3929} & \textbf{0.4228} \\ \midrule
\multirow{2}{*}{Essential Hypertension} 
& AUROC & \textbf{0.5836} & \underline{0.5622} \\
& F1    & \textbf{0.7955} & \textbf{0.7955} \\ \midrule
\multirow{2}{*}{Geriatric Depression} 
& AUROC & \textbf{0.9174} & \underline{0.8986} \\
& F1    & \textbf{0.7619} & \underline{0.7598} \\ \midrule
\multirow{2}{*}{Osteoarthritis} 
& AUROC & \underline{0.7334} & \textbf{0.7365} \\
& F1    & \underline{0.7354} & \textbf{0.7522} \\ \midrule
\multirow{2}{*}{Sleep Disturbance} 
& AUROC & \textbf{0.6243} & \underline{0.5603} \\
& F1    & \textbf{0.7473} & \underline{0.7374} \\ \midrule
\multirow{2}{*}{Sleep Efficiency} 
& AUROC & \underline{0.6174} & \textbf{0.6448} \\
& F1    & \textbf{0.7331} & \underline{0.7278} \\ \midrule
\multirow{2}{*}{\textbf{Mean}} 
& \textbf{AUROC} & \underline{0.6673} & \textbf{0.6819} \\
& \textbf{F1}    & \underline{0.6166} & \textbf{0.6435} \\ \bottomrule
\end{tabular}
\end{table}

\subsection*{Generalization Across Devices, Regions, and Diseases}
To demonstrate the robustness of \projectname as a true physical activity foundation model, we systematically evaluate its generalizability along three orthogonal dimensions: sensing devices, geographical regions, and clinical endpoints. The same three downstream datasets, NHANES 05-06~\cite{troiano2008physical}, BarKA-MS~\cite{baumer_2025_17485784}, and RESILIENT~\cite{nilforooshan_2025_16755408}, jointly cover all three dimensions, and each dataset is compared against the subset of baselines that can be reasonably adapted to its data resolution and task setup. NHANES 05-06 provides minute-level step counts and follows the same evaluation protocol as NHANES 11-14, so it is compared against the full set of foundation-model baselines reported in Section~\ref{sec:comparison}. BarKA-MS also provides minute-level step counts and is therefore compared against the general wearable foundation models (Pulse-PPG, PAT, TimeSiam, and NormWear); SSCP is excluded here because it operates on coarse-grained step sequences and cannot fully exploit the fine-grained minute-level signals that BarKA-MS provides. RESILIENT contains only hour-level coarse step counts, so minute-level foundation models cannot be applied without major modification; we therefore restrict the comparison to SSCP, which natively handles low-resolution step sequences, and evaluate our model without its minute-level micro-stream for fair comparison.

\subsubsection*{\projectname generalizes across different sensing devices.}

A fundamental bottleneck in deploying wearable models is the domain shift caused by heterogeneous hardware and sensor placement. Our model was pre-trained on wrist-worn ActiGraph accelerometer data from NHANES 11-14, while the three downstream datasets together cover three distinct sensing device types: a waist-worn ActiGraph accelerometer in NHANES 05-06, a Fitbit pedometer in BarKA-MS, and a ScanWatch pedometer in RESILIENT. We use NHANES 05-06 as a controlled probe of this generalization, since it preserves the same accelerometer modality as pre-training and isolates the placement shift from wrist to waist. As shown in Tables~\ref{tab:detailed_full} and~\ref{tab:summary_comparison}, our method achieves a mean AUROC of 0.6842 and a mean F1 of 0.2867, clearly outperforming the strongest baseline NormWear at 0.6640 and 0.2706, respectively; even using only 30\% of the training data, it still reaches an AUROC of 0.6545 and an F1 of 0.2710, surpassing most baselines trained on the full set. Our model also attains the best performance on the Fitbit and ScanWatch pedometer datasets. Overall, these results suggest that our model learns transferable biomechanical rhythms rather than overfitting to the specific hardware or sensor placement of the pre-training corpus.

\subsubsection*{\projectname generalizes across different geographical regions.}
\label{sec:cross_region}

Human physical activity patterns are heavily influenced by geographical factors, urban layouts, and cultural lifestyles. Although the pre-training corpus is drawn exclusively from a US population, the three downstream datasets jointly cover the United States (NHANES 05-06), the Swiss Confederation (BarKA-MS), and the United Kingdom (RESILIENT), enabling us to assess transfer across heterogeneous populations. We focus the quantitative comparison on the two non-US cohorts, since they simultaneously introduce different sensing devices and different region-specific lifestyles. On the Swiss BarKA-MS cohort collected with Fitbit pedometers, as shown in Table~\ref{tab:barka_ms_full}, our method reaches a mean AUROC of 0.7766 and a mean F1 of 0.8679, with its AUROC substantially exceeding the strongest baseline Pulse-PPG at 0.6773. On the UK RESILIENT cohort collected with ScanWatch pedometers, as shown in Table~\ref{tab:resilient_full}, it reaches a mean AUROC of 0.6819 and a mean F1 of 0.6435, outperforming SSCP at 0.6673 and 0.6166, respectively. This consistent advantage across two international populations indicates that our model captures fundamental behavioral phenotypes that extend beyond a single geographical or cultural context.

\subsubsection*{\projectname generalizes to novel diseases and clinical scales.}
 
Beyond geographical shifts, the external datasets introduce new clinical domains. BarKA-MS contains multiple-sclerosis-related endpoints that are entirely unseen during pre-training, while RESILIENT additionally provides psychological and cognitive endpoints assessed by professionally validated diagnostic scales detailed in Section~\ref{sec:generalizable_datasets}. This setting allows us to examine whether the learned representations capture nuanced symptomatic behaviors beyond generalized health questionnaires. On BarKA-MS, as shown in Table~\ref{tab:barka_ms_full}, \projectname's minute-level resolution captures distinctive micro-mobility signatures and reaches an AUROC of 0.8894 for MS-specific disability progression and 0.6638 for pathological fatigue, substantially outperforming all minute-level baselines. We note that Pulse-PPG attains a slightly higher mean F1 of 0.8888 against 0.8679, yet its mean AUROC is only 0.6773 compared with 0.7766. This gap reflects the strong class imbalance of BarKA-MS, where the dominant positive class inflates F1 even for models with limited discriminative ability, while the AUROC margin shows that our model is the one consistently separating positive from negative cases. On RESILIENT, as shown in Table~\ref{tab:resilient_full}, even with the minute-level micro-stream deactivated, the macro stream alone reaches an AUROC of 0.7890 on the novel Anxiety endpoint, compared with 0.6231 from SSCP, indicating that macroscopic rhythm encoding alone is sufficient to generalize to complex psychological conditions and novel clinical diagnostic scales.

\section*{Discussion}

Our findings suggest a shift in how the community conceptualizes sensing for health inference: instead of prioritizing high-frequency, high-dimensional signals, future work can explore the representational power of low-dimensional, behavior-centric data such as step counts. This shift does not imply that richer sensing modalities are unnecessary, but rather shows that simple behavioral signals may provide a strong and practical basis for broad health inference when they are modeled at scale. For researchers in ubiquitous computing and mobile health, this perspective invites a more systematic investigation of when minimal signals are sufficient, when richer sensing is required, and how different levels of sensing detail affect privacy, computation, accessibility, and clinical utility.

This opens new research questions on how minimal signals can be leveraged for maximal generalization, and what other lightweight modalities (e.g., screen usage, mobility traces) may serve as foundations for scalable health modeling. Since these signals are often passively available from commodity devices, they may support health models that can be deployed across broader populations without requiring specialized hardware or continuous collection of raw physiological data. Future work can therefore explore principled ways to evaluate the information content of such signals, identify which health outcomes are most compatible with them, and determine how signal simplicity influences robustness across datasets, devices, and demographic groups.

Additionally, \projectname highlights the importance of cross-task and cross-population learning, encouraging researchers to move beyond task-specific models toward unified frameworks that capture shared behavioral patterns across diverse health conditions. This suggests that model design should not only focus on optimizing performance for a single disease or cohort, but also on learning representations that remain useful under changes in population, disease category, data source, and deployment environment.

Another promising direction lies in integrating step-based representations with multimodal foundation models, investigating how compact signals can complement richer modalities while maintaining privacy and efficiency. Step data may provide a stable behavioral context that helps organize or regularize information from IMU signals, heart rate, sleep records, self reports, or clinical variables. Conversely, richer modalities may compensate for the limited physiological and contextual information available in step counts alone. This creates opportunities to design models that adapt the sensing burden to the task: using simple signals when they are sufficient, and adding richer signals only when they provide meaningful additional value.

\section*{Conclusion}
In this paper, we introduced \projectname, a foundation model built exclusively on step count data for broad-spectrum health risk prediction. By leveraging the ubiquity, low dimensionality, and privacy-preserving nature of step data, our approach offers a practical alternative to traditional methods that rely on high-frequency raw sensor signals. We designed a scalable pre-training framework that captures temporal dynamics and long-term behavioral patterns, enabling effective transfer across more than 20 health prediction tasks spanning diverse datasets, new populations, and novel diseases.
Our extensive experiments demonstrate that \projectname achieves strong and consistent performance across heterogeneous settings, highlighting the potential of step-based representations as a unified signal for health inference. Beyond predictive accuracy, our findings reveal interpretable and generalizable relationships between physical activity patterns and a wide range of health risks, providing new insights into activity-driven health modeling.
Overall, this work opens new opportunities for large-scale health monitoring using widely available step data.

\bibliography{sample}

\section*{Methods}

\subsection*{Framework Pipeline}
\begin{figure*}
    \centering
    \includegraphics[width=1\linewidth]{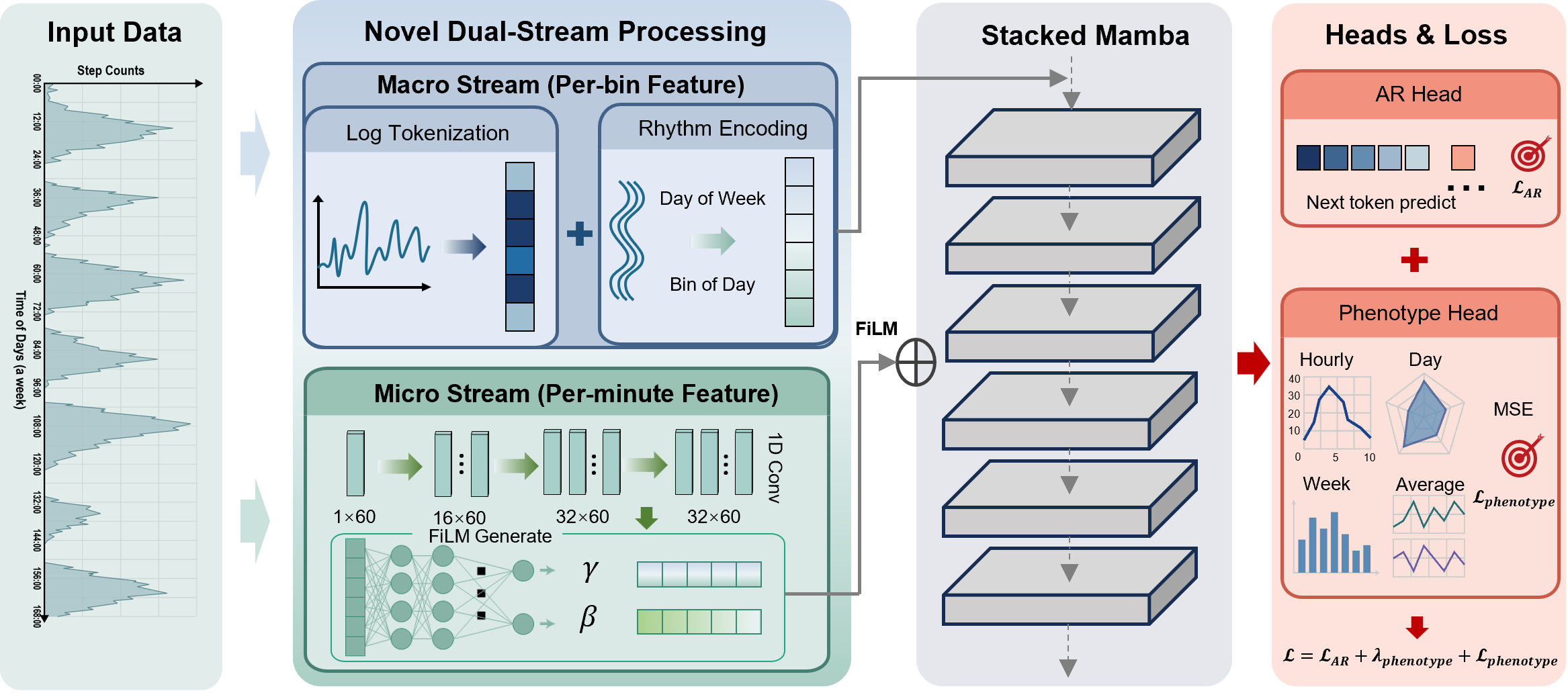} 
    \caption{\textbf{Self-supervised pre-training framework of \projectname.} Raw minute-level step counts are first encoded into hourly macro tokens using log-scaled tokenization and temporal rhythm injection, while preserving local intra-hour dynamics via a micro CNN. These dual-stream representations are fed into a Mamba backbone, dynamically modulated by FiLM of micro stream, and jointly optimized through autoregressive next-token prediction and hierarchical activity phenotype alignment.}
    \label{fig:framework}
\end{figure*}

Step-count data contain health-relevant patterns at multiple temporal scales, from within-hour activity bursts to circadian regularity and cross-day variability \cite{fujino2023decreased, hafiz2020wearable, bae2016using, low2018fitbit}. \projectname addresses this multi-scale structure through a behavior-aware self-supervised framework for pre-training. As shown in Figure~\ref{fig:framework}, it converts minute-level steps into hourly macro tokens while preserving the intra-hour waveform as local context, and uses a dual-stream Step-Mamba encoder optimized by next-token prediction and hierarchical activity-phenotype alignment. After that, the pre-trained model is fine-tuned in downstream tasks for diverse health risk prediction.

\subsection*{Behavior-Aware Pre-training}

Similar to large language models that learn representations by predicting future words from preceding context, the pre-training task learns wearable activity representations by forecasting future ambulatory patterns from historical step observations. However, step-count sequences differ from natural language in important ways: they contain strong daily and weekly periodicity, long sparse sedentary intervals, and multi-scale behavioral phenotypes that may be more clinically informative than any single next-step transition. Therefore, our pre-training design goes beyond standard next-token prediction by aligning the latent space with interpretable physical activity statistics at hourly, daily, and weekly levels.

\subsubsection*{Data Preparation and Tokenization}

We pre-train the model on large-scale, unlabeled NHANES 2011--2014 step count data \cite{nhanesdata}. For each participant, seven weekday records are concatenated by participant identifier to construct a continuous Sunday-to-Saturday temporal sequence. This yields a raw minute-level sequence 
\[
\mathbf{x}_i = [x_{i,1}, x_{i,2}, \ldots, x_{i,10080}],
\]
where each element denotes the step volume recorded within a specific minute.

To seamlessly model both the macroscopic weekly trends and the microscopic activity morphology, we partition the sequence into hourly bins while retaining the raw minute-level signals within each bin. Given a bin width of \(W=60\) minutes, each 7-day sequence yields \(N=168\) hourly bins. The aggregated step volume of the \(t\)-th hourly bin is computed as:
\[
s_{i,t} = \sum_{m=1}^{W} x_{i,(t-1)W+m}.
\]

The tokenizer serves as a critical signal-processing bottleneck that discretizes continuous measurements into discrete vocabulary IDs prior to sequence encoding. This design is tailored to the physiological nature of step data. While step counts are an established proxy for ambulatory health \cite{bassett2010pedometer,tudor2004descriptive}, they possess a highly skewed, non-Gaussian distribution. Extensive sedentary periods produce an abundance of near-zero values, whereas transient intense exercises generate occasional extreme spikes. A naive linear discretization would waste vocabulary capacity on rare high-count outliers while failing to resolve subtle, clinically meaningful variations within the low-to-moderate activity spectrum.

To mitigate this, we propose a log-scaled tokenizer with a vocabulary size of \(V=256\). Let \(S_{\max}=6000\) denote the physiological saturation threshold and let \(q(s)\) be the token assigned to an hourly step sum \(s\). The tokenizer is formulated as:
\[
q(s) =
\begin{cases}
\mathrm{round}\!\left(
\dfrac{\log(1+s)}{\log(1+S_{\max})}(V-2)
\right), & 0 \leq s < S_{\max},\\[8pt]
V-1, & s \geq S_{\max}.
\end{cases}
\]
The logarithmic transformation maps \(s\) to a normalized space in \([0,1)\), ensuring that the model allocates higher resolution to the low-moderate activity ranges where nuanced behavioral shifts occur. The overflow token \(V-1\) robustly compresses all extreme values (\(\geq S_{\max}\)) into a single high-intensity category. This mechanism guarantees a compact and highly representative input sequence, which is essential for scalable wearable foundation models \cite{narayanswamy2024scaling,qiu2025towards}.

\subsubsection*{Temporal Rhythm Encoding}

Human physiology and behaviors are intrinsically governed by temporal rhythms, such as diurnal circadian cycles and weekday-versus-weekend routines. Consequently, an identical step volume can carry vastly different behavioral implications depending on its temporal context (e.g., a 2,000-step walk at noon indicates a regular commute, whereas the same volume at 3:00 AM may indicate severe sleep disturbance). To explicitly ground the sequence in physical time, we inject deterministic periodic features into the binned representations, inspired by time-space encoding strategies \cite{vaswani2017attention,kazemi2019time2vec}.

Let \(B=24\) be the number of hourly bins per day. For the \(t\)-th hourly bin, its bin-of-day index \(b_t\) and day-of-week index \(d_t\) are defined as:
\[
b_t = (t-1) \bmod B,
\qquad
d_t = \left\lfloor \frac{t-1}{B} \right\rfloor.
\]
For a periodic index \(a\) with period \(P\), we construct multi-frequency Fourier features as:
\[
\phi_P(a) = \left[ \sin\!\left(\frac{2\pi k a}{P}\right), \cos\!\left(\frac{2\pi k a}{P}\right) \right]_{k=1}^{K_P}.
\]
The diurnal cycle is encoded via \(\phi_B(b_t)\), while the weekly rhythm is encoded via \(\phi_7(d_t)\). We concatenate these Fourier features and map them to the activity representation dimension \(D\) with a lightweight multi-layer perceptron (MLP):
\[
\mathbf{r}_t =
\mathrm{MLP}\!\left(
\left[\phi_B(b_t); \phi_7(d_t)\right]
\right) \in \mathbb{R}^{D}.
\]
This continuous encoding smoothly captures temporal proximities (e.g., the seamless transition across the midnight boundary). We inject the resulting temporal code into the initial activity representation \(\mathbf{z}_t\):
\[
\tilde{\mathbf{z}}_t = \mathbf{z}_t + \mathbf{r}_t.
\]
This temporal grounding enables the model to effortlessly differentiate context-dependent behaviors that share similar quantitative volumes.
\subsubsection*{Dual-stream Step-Mamba Encoder}

A major challenge in multi-day wearable sensing is reconciling macro-level long-range trends with micro-level instantaneous bursts. We address this using a Mamba-based architecture \cite{gu2023mamba} named dual-stream Step-Mamba. 

The \textbf{Macro Stream} maps each hourly step token to a 256-dimensional embedding and models the week-long trajectory using a Mamba backbone. The choice of Mamba over conventional Transformers is deliberate: its selective state-space design provides linear-time complexity, making it exceptionally efficient for capturing ultra-long-range dependencies inherent in multi-day tracking \cite{gu2023mamba}.

Concurrently, the \textbf{Micro Stream} processes the raw 60-minute waveform within each corresponding hourly bin using a lightweight 1D convolutional network. This stream acts as a feature extractor for local morphology (e.g., distinguishing a sustained 30-minute jog from sporadic household movements that yield the same hourly sum). 

Rather than naively concatenating the micro features, which could dilute the temporal sequence, we use Feature-wise Linear Modulation (FiLM) to dynamically condition the macro hidden states. Specifically, at selected intermediate layers, the micro-stream cues are injected as:
\[
\tilde{\mathbf{h}}_t = \mathrm{LN}\left(\mathbf{h}_t \odot \left(1 + \alpha \tanh(\gamma_t)\right) + \alpha \tanh(\beta_t)\right),
\]
where the scaling factor \(\gamma_t\) and shifting factor \(\beta_t\) are generated entirely from the local micro-stream features, \(\alpha\) bounds the modulation strength, and \(\mathrm{LN}\) denotes layer normalization. Through this mechanism, minute-level morphological details gracefully "guide" the macro sequence representation without disrupting the global autoregressive flow.

\subsubsection*{Hierarchical Activity Phenotype Alignment}

To optimize the dual-stream backbone, we use two complementary self-supervised objectives. The primary objective is the \textbf{Autoregressive Next-Token Prediction}, which enforces strict temporal causality. Given the hidden states \(\mathbf{X}_{1:t}\), a linear language-modeling head predicts the token for the next bin, optimized via cross-entropy loss:
\[
\mathcal{L}_{\mathrm{AR}} = -\sum_{t=1}^{N-1} \log p(y_{t+1} \mid \mathbf{X}_{1:t}).
\]

However, optimizing solely for local next-step forecasting may cause the model to under-represent macro-level clinical phenotypes, such as cross-day regularity and total sedentary burden. To bridge this gap, we introduce a \textbf{Hierarchical Phenotype Alignment} objective. The phenotype targets are extracted directly from the raw step sequence and z-score normalized, encouraging the model to encode health-relevant indicators across three temporal scales: hourly statistics such as log step volume, active/sedentary ratios, and step entropy; daily aggregate statistics obtained from pooled hourly embeddings; and weekly phenotypes that summarize cross-day consistency, variability, and total activity burden.

The composite pre-training loss is therefore written as
\[
\mathcal{L}
=
\mathcal{L}_{\mathrm{AR}}
+
\lambda_{\mathrm{phenotype}}\mathcal{L}_{\mathrm{phenotype}},
\]
where \(\lambda_{\mathrm{phenotype}}\) controls the overall contribution of phenotype alignment. The phenotype loss itself is defined without additional internal weighting:
\[
\mathcal{L}_{\mathrm{phenotype}}
=
\mathcal{L}_{\mathrm{hour}}
+
\mathcal{L}_{\mathrm{day}}
+
\mathcal{L}_{\mathrm{week}}
.
\]
All phenotype losses are computed as mean-squared errors on z-score-normalized targets.

\subsection*{Downstream Adaptation: Disease Risk Prediction}

Following pre-training, the Step-Mamba encoder is evaluated on 21 diverse downstream disease prediction tasks. To demonstrate the intrinsic generalizability of the learned representations, we adopt a strict MLP evaluation protocol. 

The pre-trained dual-stream backbone is completely frozen, serving purely as a universal physiological feature extractor. For a given participant's 7-day sequence, the backbone outputs a high-dimensional contextualized embedding. Only a simple task-specific Multilayer Perceptron (MLP) head is attached and fine-tuned using the supervised disease labels from downstream datasets such as disease data in NHANES \cite{nhanesdata}. This decoupling allows the model to leverage massive unannotated datasets to build a robust foundation, while maintaining computational efficiency and avoiding over-fitting during specialized clinical adaptations.

% *(Note: In the final manuscript, this subsection will detail the definitions of the 24 disease endpoints, label curation, train/validation/test splits, strategies for mitigating class imbalance, and specific evaluation metrics such as AUROC and AUPRC.)*

\subsection*{Implementation Details}

The core dual-stream autoregressive Step-Mamba model comprises approximately 3.4 million trainable parameters. The architecture features 6 Mamba layers with a hidden dimension of 256, a state dimension of 16, a convolution width of 4, and an expansion factor of 2. The micro stream uses four 1D convolutional layers with channel dimensions (1, 16, 32, 32), and FiLM modulation from its outputs is applied immediately before Mamba layers 0, 2 and 4.

The pre-training phase was executed over 200 epochs with a batch size of 128. We utilized the AdamW optimizer with a weight decay of 0.05 and momentum betas of \((0.9, 0.95)\). The learning rate followed a 2-epoch warmup to a peak of \(3\times10^{-4}\), decaying to a minimum of \(1\times10^{-5}\), with gradient clipping enforced at 1.0. To optimize computational throughput, mixed-precision training (FP16) was employed. All experiments were conducted on a workstation equipped with NVIDIA RTX 4090 GPUs. 

% *(Note: The final manuscript will report the exact number of GPUs, CUDA version, and approximate total wall-clock time for pre-training.)

\subsection*{Dataset}

\subsubsection*{Pre-Training Datasets}
We used the minute-level step count dataset from NHANES 2011--2014 \cite{nhanesdata} as our pre-training corpus without health label. In these NHANES cycles, participants wore an ActiGraph GT3X+ accelerometer on the non-dominant wrist for seven consecutive days. The original NHANES accelerometer data were not provided as direct step counts, but as raw triaxial accelerometer signals recorded at 80 Hz. Therefore, these raw signals need to be processed by step counting algorithms before they can be used as step count sequences. Koffman and Muschelli processed the raw NHANES accelerometer data with several step counting algorithms and released minute-level step count and physical activity variables for 14,693 individuals on PhysioNet \cite{PhysioNet-minute-level-step-count-nhanes-1.0.1}. In this study, we specifically used the scrfsteps data from this release, which correspond to step count estimates generated by the random forest version of the Stepcount algorithm. We organized each participant's records into one week, 24 hour step count sequences. After filtering participants with available step count and disease records, the pre-training set contained approximately 14,000 individuals and 141.12 million minute-level step count observations. This pre-training dataset enables the model to learn generalizable representations from large scale, low dimensional behavioral signals before being transferred to downstream health inference tasks.

\subsubsection*{Fine-Tuning Datasets for Downstream Health Prediction}

% We further linked the processed step count data with the original NHANES health questionnaires using SEQN as the participant identifier. Disease labels were derived from four questionnaire modules: Medical Conditions Questionnaire (MCQ), Blood Pressure \& Cholesterol Questionnaire (BPQ), Diabetes Questionnaire (DIQ), and Mental Health - Depression Screener Questionnaire (DPQ). These modules provide self reported or questionnaire derived information on medical conditions, blood pressure and cholesterol, diabetes, and depressive symptoms. In total, we constructed labels for 21 health outcomes: anemia, angina, arthritis, cancer, chronic bronchitis, gout, coronary heart disease, emphysema, heart attack, heart failure, kidney weak/failing, overweight, stroke, thyroid problems, vision trouble, memory cognitive function difficulty, diabetes, high blood pressure, high cholesterol, sleep disorder, and depression. 

\begin{figure}[t]
    \centering
    \includegraphics[width=\linewidth]{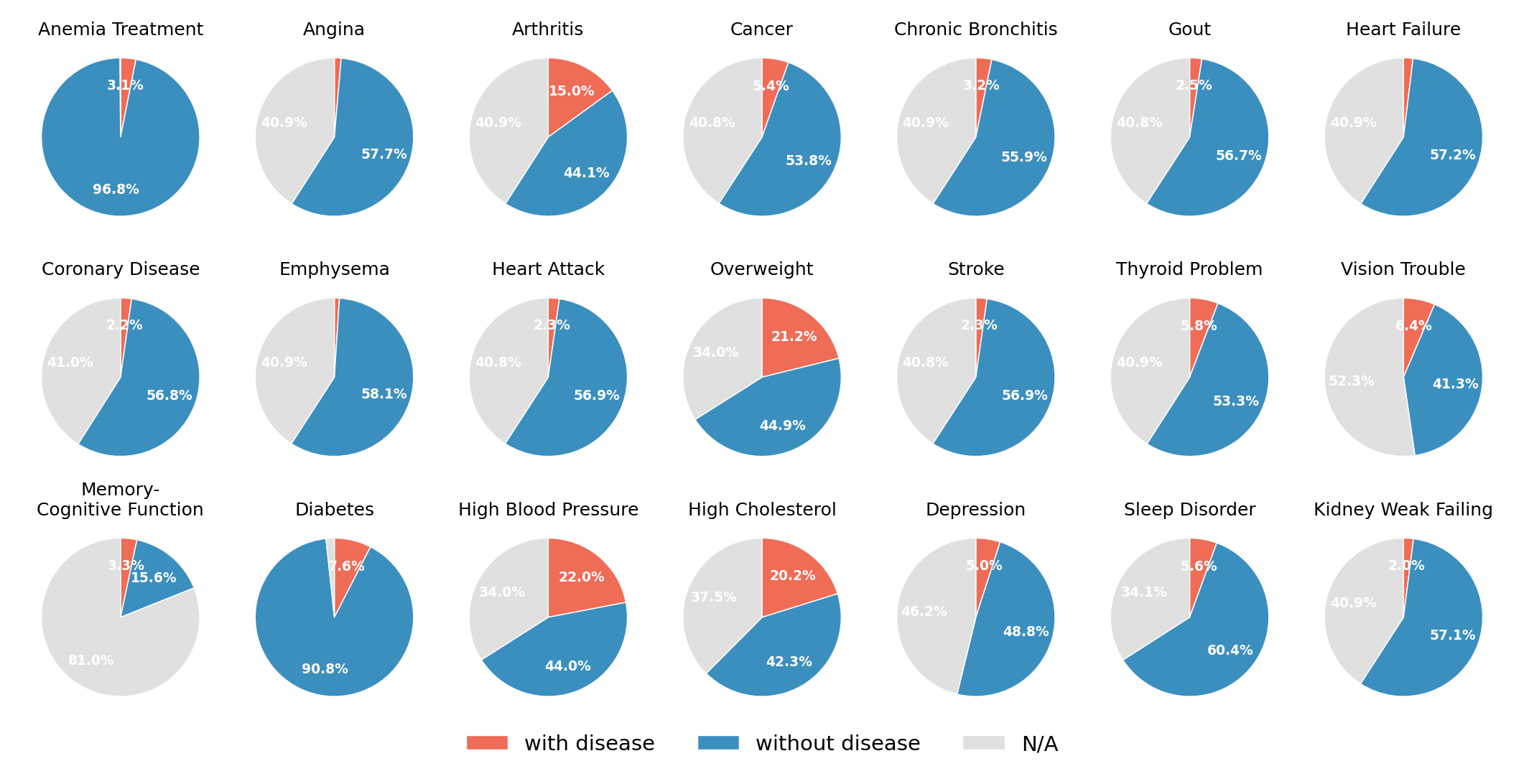}
    \caption{\textbf{Disease prevalence distribution across the NHANES 2011--2014 dataset.} Each row corresponds to a health outcome, where the three color blocks denote the proportions of positive cases, negative cases, and missing, respectively.}
    \label{fig:disease}
\end{figure}

We further linked the processed step count data with the original NHANES health questionnaires using SEQN as the participant identifier. Health outcome labels were mainly derived from four questionnaire modules: the Medical Conditions Questionnaire (MCQ), Blood Pressure \& Cholesterol Questionnaire (BPQ), Diabetes Questionnaire (DIQ), and Mental Health - Depression Screener Questionnaire (DPQ). These modules provide a comprehensive set of self-reported and questionnaire-derived physiological and psychological indicators as shown in Figure~\ref{fig:disease}. To facilitate a structured evaluation, we categorized the 21 evaluated health outcomes into four broad clinical groups: \textit{Cardiovascular \& Circulatory} (angina, heart failure, coronary heart disease, heart attack, stroke, high blood pressure); \textit{Metabolic, Endocrine \& Renal} (diabetes, overweight, high cholesterol, gout, thyroid problem, kidney weak/failing); \textit{Respiratory \& Systemic Conditions} (emphysema, chronic bronchitis, arthritis, cancer, anemia treatment); and \textit{Neurological, Mental \& Sensory} (memory/cognitive difficulties, depression, sleep disorder, vision trouble), as reported in Table~\ref{tab:comparison_auroc}.

\subsubsection*{Generalizable Downstream Datasets}
\label{sec:generalizable_datasets}
% To evaluate whether Prancer learns generalizable step-based representations beyond the NHANES cohort, we evaluate it on five external downstream datasets with different health-related prediction tasks, as summarized in Table~\ref{tab:downstream_datasets}. These datasets span heterogeneous populations, countries or regions, temporal resolutions, observation durations, and disease categories. Specifically, the downstream datasets include hourly and minute-level step-count records collected from the United Kingdom, the Swiss Confederation, the United States, and South Korea, with observation windows ranging from seven days to over 100 days. The datasets also cover different sensing devices, including pedometers from \textbf{ScanWatch}, \textbf{Fitbit}, \textbf{smartwatches}, and \textbf{ActiGraph accelerometers}. This evaluation setup allows us to examine whether a foundation model trained on large scale step sequences can transfer across different devices, datasets, populations, and health outcomes.

To evaluate whether \projectname learns generalizable step-based representations beyond the NHANES cohort, we evaluate it on three downstream datasets with different health-related prediction tasks, as summarized in Table~\ref{tab:downstream_datasets}. These datasets span heterogeneous populations, countries or regions, temporal resolutions, observation durations, and disease categories. Specifically, the downstream datasets include hourly and minute-level step-count records collected from the United Kingdom, the Swiss Confederation, and the United States, with observation windows ranging from seven days to over 100 days. The datasets also cover different sensing devices, including pedometers from \textbf{ScanWatch}, \textbf{Fitbit}, and \textbf{ActiGraph accelerometers}. This evaluation setup allows us to examine whether a foundation model trained on large-scale step sequences can transfer across sensing devices, geographical regions, and diseases.

Additionally, each downstream dataset is associated with health-related tasks that reflect its study population and data collection context. For RESILIENT, several labels were originally provided as multi-level diagnostic scales, which we converted into binary prediction tasks to match our downstream evaluation protocol
~\cite{spitzer2006brief}: anxiety was defined as GAD-7 \(\geq 10\), depression as PHQ-9 \(\geq 10\), geriatric depression as GDS \(\geq 5\), cognitive impairment as ACE \(\leq 82\), sleep disturbance as WASO \(\geq 90\) minutes, and low sleep efficiency as sleep efficiency \(<0.85\), while essential hypertension and osteoarthritis were retained as original labels. BarKA-MS provides fatigue and disability endpoints, and NHANES 2005-2006 follows the same label definitions as NHANES 2011--2014 except that gout and memory-cognitive function are unavailable, yielding 19 shared health outcomes. Although many of these conditions are not conventionally inferred from step-count sequences alone, we include them to evaluate the versatility of \projectname in learning meaningful behavioral representations from low-dimensional physical activity data.

\begin{table*}[t]
    \centering
    \caption{\textbf{Downstream datasets and tasks.} Summary of the three external datasets used to evaluate the generalizability of \projectname. All listed tasks are formulated as binary classification of health outcomes derived from each dataset's clinical or self-reported labels.}
    \label{tab:downstream_datasets}
    \resizebox{\textwidth}{!}{
    \begin{tabular}{l c c c}
        \toprule
            & RESILIENT~\cite{nilforooshan_2025_16755408} 
            & BarKA-MS~\cite{baumer_2025_17485784} 
            & NHANES 2005--2006~\cite{troiano2008physical}\\
        \midrule
        Sampling Interval & 1 hour & 1 minute & 1 minute\\
        Num. of Participants & 72 & 44 & 7,455\\
        % Num. of Samples & 1345 & 280 & \\
        Time per Participant & Over 100 days & 9 weeks & 7 days \\
        Sensor Type & Pedometer (ScanWatch) & Pedometer (Fitbit) & Accelerometer (ActiGraph)\\
        Country or Region & the United Kingdom & the Swiss Confederation & the United States \\
        % Ethnicity Included & NAN & NAN & several ethnicity groups\\
        \midrule
        Tasks Available
        & \begin{tabular}{@{}c@{}}
            Essential Hypertension\\ 
            Osteoarthritis\\ 
            Anxiety\\ 
            Depression\\ 
            Geriatric Depression\\
            Cognitive Impairment\\
            Sleep Disturbance\\ 
            Sleep Efficiency
          \end{tabular}
        & \begin{tabular}{@{}c@{}}
            Fatigue\\ 
            Disability
          \end{tabular}
        & \begin{tabular}{@{}c@{}}
            19 shared NHANES2011-2014 diseases\\
            excluding Gout and\\
            Memory-Cognitive Function
          \end{tabular}\\
        \bottomrule
    \end{tabular}
    }
\end{table*}

\end{document}